\documentclass[12pt,a4paper]{article}

\usepackage{graphicx}
\usepackage{pdfpages}
\usepackage{geometry}
\usepackage{fancyhdr}
\usepackage{amsmath}
\usepackage{amssymb}
\usepackage{color}
\usepackage{subcaption}
\usepackage{multirow}
\usepackage{booktabs}
\usepackage{tikz}

\title{An Efficient Procedure for Computing \\Bayesian Network Structure Learning}
\author{Hongming Huang and Joe Suzuki \\ Osaka University}
\date{}

\begin{document}

\maketitle

\begin{abstract}
We propose a globally optimal Bayesian network structure discovery algorithm based on a progressively leveled scoring approach. Bayesian network structure discovery is a fundamental yet NP-hard problem in the field of probabilistic graphical models, and as the number of variables increases, memory usage grows exponentially. The simple and effective method proposed by Silander and Myllymäki has been widely applied in this field, as it incrementally calculates local scores to achieve global optimality. However, existing methods that utilize disk storage, while capable of handling networks with a larger number of variables, introduce issues such as latency, fragmentation, and additional overhead associated with disk I/O operations. To avoid these problems, we explore how to further enhance computational efficiency and reduce peak memory usage using only memory. We introduce an efficient hierarchical computation method that requires only a single traversal of all local structures, retaining only the data and information necessary for the current computation, thereby improving efficiency and significantly reducing memory requirements. Experimental results indicate that our method, when using only memory, not only reduces peak memory usage but also improves computational efficiency compared to existing methods, demonstrating good scalability for handling larger networks and exhibiting stable experimental results. Ultimately, we successfully achieved the processing of a Bayesian network with 28 variables using only memory.
\end{abstract}


\section{Introduction}
According to Judea Pearl's definition (Pearl, 1988), a Bayesian Network (BN) is a probabilistic graphical model that uses a directed acyclic graph (DAG) to represent a set of random variables and their conditional dependencies. Each node in the network represents a random variable, and each directed edge represents a direct dependency or causal relationship between the variables. Specifically, a Bayesian Network consists of three main components: 1. Nodes (Variables): These represent the random variables, which can be either discrete or continuous. 2. Directed Edges: These indicate direct causal or dependency relationships between variables. The structure of the DAG ensures that there are no cycles in the network. 3. Conditional Probability Distributions (CPDs): For each node, given the values of its parent nodes, the CPD defines the probability distribution of the node's values.

A key feature of Bayesian Networks is their ability to leverage conditional independence to decompose the joint probability distribution, thereby making computation and inference more efficient. Specifically, if each variable \( x_i \) in a set of variables \( X \) depends only on its parent set \( \pi_i \), then the joint probability distribution \( P(X) \) can be expressed as:
\[ P(X) = \prod_{i} P(x_i \mid \pi_i) \]

In a Bayesian Network, if two variables \( X \) and \( Z \) are conditionally independent given another variable \( Y \), it can be formally expressed as: 
$X \mathrel{\perp\!\!\!\perp} Z \mid Y$.

This means that, given \( Y \), the joint probability distribution of \( X \) and \( Z \) can be factored into the product of their individual conditional probability distributions:
$P(X, Z \mid Y) = P(X \mid Y) \cdot P(Z \mid Y)$. As shown in Figure 1, the joint probability distributions for the three cases are \( P(Y) P(X \mid Y) P(Z \mid Y) \) for (a), \( P(X) P(Y \mid X) P(Z \mid Y) \) for (b), and \( P(Z) P(Y \mid Z) P(X \mid Y) \) for (c), respectively. However, they can also all be represented by \( P(X, Y) P(Y, Z) / P(Y) \). This means that these three Bayesian Networks are Markov equivalent. In this paper, we will adhere to Markov equivalence and will not distinguish between them.
\begin{figure}[h!]
    \centering
    \includegraphics[width=0.8\textwidth]{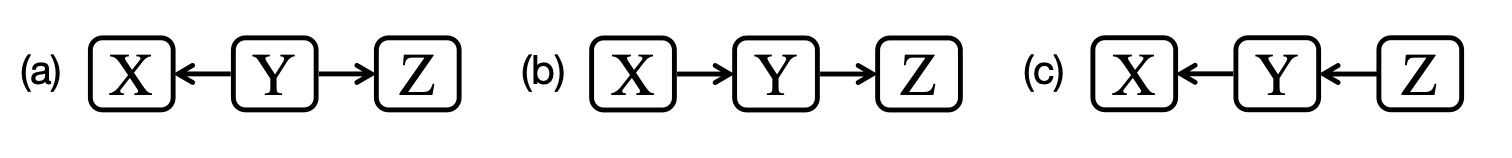}
    \caption{Three Markov equivalent Bayesian Networks.}
\end{figure}

Structure learning algorithms can generally be categorized into three types: score-based learning, constraint-based learning, and hybrid methods. The first type is constraint-based methods, which typically use statistical tests to identify conditional independence relationships from data, thus eliminating and orienting edges. For instance, the PC algorithm (Spirtes and Glymour, 1991) incrementally increases the size of the conditioning set to check the independence relationship between each pair of nodes. After finding the separation set, it removes the corresponding edges. Under the assumptions of faithfulness and causal sufficiency, it only considers adjacent nodes of a node as the conditioning set, reducing the number and complexity of conditional independence tests. The PC-Stable algorithm (Colombo and Maathuis, 2014) improves upon the PC algorithm by recalculating adjacency relationships before processing all conditional independence tests of each conditioning set size, preventing erroneous edge deletions. By considering all possible separation sets to determine V-structures, it reduces sensitivity to node order and marks conflicting edges with bidirectional arrows, avoiding arbitrary edge orientation issues. Constraint-based methods largely rely on local statistical test results, allowing them to scale to large datasets. However, they are sensitive to the accuracy of statistical tests and may perform poorly with insufficient or noisy data.

The second type is score-based methods, representing traditional machine learning approaches, which search different graph structures to maximize an objective function (Cooper \& Herskovits, 1992; Heckerman, 1998). These methods consist of two elements: (a) an objective function for evaluating each explored graph in the search space, and (b) a search strategy for determining which path to explore in the possible graph space. The main challenge of score-based learning is to find a high-scoring or ideally the highest-scoring graph, striving to find a global optimum among extensive local possibilities.

We first describe the objective function, as it applies to all score-based algorithms. Objective functions are divided into two categories: Bayesian scores and information-theoretic scores. Bayesian scores primarily focus on goodness-of-fit and allow the incorporation of prior knowledge. Bayesian scores define the posterior probability distribution of a network structure based on data, with the structure having the highest posterior probability being considered the optimal structure. A representative of these scoring functions is the Bayesian Dirichlet equivalent uniform (BDeu) score (Buntine, 1991), which uses the Dirichlet distribution as a prior distribution and assumes all prior parameters are equal, thus satisfying Markov equivalence. However, BDeu also has its limitations. In Suzuki's discussion on the regularity of scoring rules (Suzuki, 2017), he provides an example showing that BDeu is actually not regular. Simply put, assuming there are three variables \(X\), \(Y\), and \(Z\), when variable \(X\) can be sufficiently explained by \(Y\), BDeu tends to choose \(\{Y, Z\}\) as the parent set of \(X\). At this point, the model contains unnecessary complexity. Conversely, information-theoretic scores consider goodness-of-fit while explicitly considering model complexity to avoid overfitting. Fit is measured by the likelihood of the structure given the data or the amount of information compressed from the data to the structure. Scoring functions in this category include Minimum Description Length (MDL) (Suzuki, 1996), Akaike Information Criterion (AIC) (Akaike, 1973), and Mutual Information Test (MIT) (de Campos, 2006), among others. All these scoring functions are decomposable, meaning during the search process, only the scores of nodes affected by graph changes need to be recalculated, significantly improving computational efficiency (Heckerman, 1998).

Next is the search strategy, which can be divided into local search strategies and global search strategies. Score-based learning has been proven to be NP-hard (Chickering, 1996), and due to its complexity, early research mainly focused on developing approximate algorithms as local search strategies. Bouckaert (1994) removed the restriction of predefined node order and described a generic greedy hill-climbing (HC) search algorithm on the space of DAGs. This may be the simplest and most commonly used search strategy. The starting point of HC search can be any DAG, such as a randomly generated one, one generated by another structure learning algorithm, or even one based on expert knowledge. However, it usually starts from an empty graph. HC is a very efficient algorithm but may get stuck in a poor-scoring local maximum. To escape local maxima, several techniques have been adopted. Heckerman et al. (1995) used local restarts, where the DAG is randomly perturbed at a local maximum and hill climbing is restarted from the perturbed DAG. Bouckaert (1995) used a tabu list to prevent returning to recently visited DAGs and allowed some score-reducing changes to the graph.

Increasingly more global search strategies are emerging, such as dynamic programming, which starts with single variables and constructs optimal sub-networks for increasingly larger variable sets until the global optimal network is found (Silander \& Myllymäki, 2012; Singh \& Moore, 2005). The globally optimal Bayesian network structure learning algorithm proposed by Silander and Myllymäki (2012) is very intuitive, easy to understand, and code. It first calculates local scores, then efficiently solves the optimal parent set for each variable through dynamic programming, and then solves the sink node for each subset. A sink node is a node in a Directed Acyclic Graph (DAG) that has no child nodes. For each local structure, there must be a sink node. After this, we can derive the globally optimal order from the sink nodes and combine the optimal parent set information to obtain the globally optimal Bayesian network. A branch and bound algorithm (BB) was proposed for learning Bayesian networks (Campos and Ji, 2011). This algorithm first creates a cyclic graph by allowing each variable to obtain the optimal parent node from all other variables. Then, it uses a best-first search strategy to break cycles one at a time. The algorithm uses approximate algorithms to estimate initial upper-bound solutions for pruning. The algorithm occasionally expands the worst node in the search frontier, hoping to find a better network to update the upper bound. When complete, the optimal network structure found by the algorithm is a subgraph of the initial cyclic graph. If the algorithm exhausts memory before finding a solution, it switches to a depth-first search strategy to find a suboptimal solution.

The third type is hybrid methods that combine score-based and constraint-based approaches, attempting to combine the best features of each method. Perhaps the most common combination method is to use constraint-based methods to restrict the search space, then use score-based methods within that space to find the locally or globally highest-scoring graph. A hybrid MCMC algorithm (Kuipers et al., 2022) initially employs the PC constraint-based algorithm to establish a restricted search space, followed by MCMC sampling in the node ordering space or partition space. The sampling occurs within an initial space identified by the PC algorithm, but each node is permitted an additional parent outside the initial space, thereby relaxing the restriction.

In this paper, we consider the second type, namely score-based learning methods. We choose the quotient Jeffreys' score (Suzuki, 2017), which has been proven to be both regular and satisfies Markov equivalence, in contrast to BDeu. The specific content will be detailed in Section 2. The algorithm proposed in this paper further improves upon the algorithm by Silander and Myllymäki (2012). To address the NP-hard problem of Bayesian networks, they opted to use disk storage for temporarily unnecessary variables. This effectively reduces peak memory usage, but the problem is that disk read/write speed is much lower than memory. Therefore, frequent writing to and reading from the disk significantly reduces overall execution efficiency. Long and intensive disk I/O operations may cause unexpected system issues, affecting the algorithm's stability and reliability. To avoid risks associated with disk usage, we choose to use only memory to obtain the globally optimal Bayesian network. In subsequent experiments, we will also provide the stability results of the algorithm when using memory only. However, it is foreseeable that not using disk storage will reduce the maximum number of variables that can be supported. Therefore, in this paper, we propose a tiered algorithm, where each tier's calculation only requires information and data from the previous tier, reducing peak memory usage to achieve Bayesian networks with more variables. This algorithm only needs to traverse all local structures once, so even if the theoretical time complexity is similar to existing methods, there will be significant improvements. We will conduct experimental comparisons of the two methods under memory-only conditions.

The structure of this paper is as follows: Section 2 introduces the background knowledge, with Sections 2.1, 2.2, and 2.3 explaining Bayesian Network Structure Learning (BNSL), marginal likelihood, and the quotient Jeffreys’ score, respectively. Section 3 reviews the existing work. Section 4 presents our method. Section 5 demonstrates the experimental results. Section 6 concludes with a discussion and future work directions.

\section{Background}
\subsection{Bayesian Network Structure Learning}
Suppose that given \( n \) tuples of examples
\[
\begin{aligned}
X^{(1)} &= x_{i,1}, & X^{(2)} &= x_{i,2}, & \ldots, & & X^{(p)} &= x_{i,p},
\end{aligned}
\]
where \( i = 1, 2, \ldots, n \), with respect to \( p \) variables \( X^{(1)}, X^{(2)}, \ldots, X^{(p)} \). 

A fundamental step in structure learning is finding the optimal parent sets for each node within the network. The joint probability of the variables \( X^{(1)}, \ldots, X^{(p)} \) in a BN is factorized according to the network structure as follows:
\begin{equation}
P(X^{(1)}, \ldots, X^{(p)}) = P(X_{1})P(X_{2}|\pi_{2})\cdots P(X_{i}|\pi_{i})\cdots P(X_{p}|\pi_{p}) = \prod_{i=1}^{p} P(X_i \,|\, \pi_{i}),
\end{equation}
where \( X_i \) denotes the vertex at i-th position in the structure and \( \pi_{i} \) denotes the parent set of the vertex \( X_i \). 
We identify a BN structure that has the maximum posterior probability, given the \( p \) variables and the prior probabilities over structures and parameters. Specifically, maximizing posterior probability is achieved by considering marginal likelihood. 

\subsection{Marginal Likelihood}
In Bayesian inference, the posterior probability \( P(G | D) \) represents the probability of a BN structure \( G \) given the data \( D \). According to Bayes' theorem, the posterior probability can be expressed as:

\begin{equation}
P(G | D) = \frac{P(D | G) P(G)}{P(D)}.
\end{equation}

Here, \( P(D | G) \) is the marginal likelihood (also known as model evidence), representing the probability of the data \( D \) given the network structure \( G \). \( P(G) \) is the prior probability, representing the prior knowledge about the network structure \( G \) before observing the data \( D \). \( P(D) \) is a normalization constant, representing the total probability of the data \( D \).

In the structure learning process, our goal is to maximize the posterior probability \( P(G | D) \) to find the most likely network structure \( G \). However, since \( P(D) \) is constant for all candidate structures \( G \), we can ignore it when comparing different network structures. Maximizing the posterior probability is equivalent to maximizing \( P(D | G) P(G) \).

\( P(D | G) \) is the marginal likelihood, which considers the joint probability of all possible parameters \( \theta \). And we compute it as

\begin{equation}
P(D | G) := \int_\Theta P(D | G, \theta) P(\theta | G) d\theta
\end{equation}

This means that we not only consider the likelihood of the data given the network structure \( P(D | G, \theta) \), but also integrate the uncertainty of the parameters and the prior distribution \( P(\theta | G) \). Therefore, the process of maximizing the posterior probability \( P(G | D) \) is actually achieved by maximizing the marginal likelihood \( P(D | G) \) while also considering the prior probability \( P(G) \). 

We then use \( Q(X) \) to represent the marginal likelihood \( P(D | G) \). More specifically, for a set of realized values \( \mathcal{X} \) for a random variable \( X \), and a parameter set \( \Theta \subseteq \mathbb{R}^d \). Given a sample size \( n \) and i.i.d. observations \( x_1, \ldots, x_n \in \mathcal{X} \), the model is described by a family of probability distributions \( \{p(\cdot|\theta)\}_{\theta\in \Theta} \) and a prior distribution \( \varphi \) over \( \Theta \). The marginal likelihood is defined as
\begin{equation}
Q(X) := \int_\Theta \varphi(\theta) \prod_{i=1}^n p(x_i|\theta) d\theta,
\end{equation}
which synthesizes the likelihood across all parameter values, weighted by the prior. According to Equation (1), setting \( V = \{X^{(1)}, \ldots, X^{(p)}\} \), the maximum posterior probability for identifying a BN structure can be expressed as follows
\begin{equation}
R(V) := 
 \max_{\pi_{i}}\prod_{i=1}^{p} \frac{Q(X_{i}, \pi_{i})}{Q(\pi_{i})}, \hspace{3mm}\pi_{i} \subseteq V\backslash \{ X_{p}, \cdots,X_{i} \}.
\end{equation}

\subsection{Quotient Jeffreys’ score}
In this paper, we focus on multivariate and complete discrete data, and each variable has a finite number of values. According to the knowledge about marginal likelihood and Jeffreys’ prior  (Jeffreys, 1939; Krichevsky and Trofimov, 1981), the marginal likelihood \( Q(X) \) shown above can be simplified as
\begin{equation}
Q(X) := \prod_{i=1}^n \frac{c_{i-1}(x_i) + 0.5}{i - 1 + 0.5\sigma(X)},
\end{equation}
where \(c_{i-1}(x)\) is the number of occurrences of \(x\) in \((x_1, \cdots, x_{i-1})\) which starts from 0, and \( \sigma(X)\) equals to the number of different values that \( X \) takes.

When computing the parent set, we apply quotient Jeffreys’ score
\begin{equation}
 Q(X|Y) = \frac{Q(X, Y)}{Q(Y)}.
\end{equation}
For example, if \( (x_1, \ldots, x_5) = (0, 1, 0, 1, 1) \) and \( (y_1, \ldots, y_5) = (0, 0, 1, 1, 1) \), then we have 
\[
Q(X) = \frac{0 + 0.5}{0 + 0.5\cdot2} \cdot \frac{0 + 0.5}{1 + 0.5\cdot2} \cdot \frac{1 + 0.5}{2 + 0.5\cdot2} \cdot \frac{1 + 0.5}{3 + 0.5\cdot2} \cdot \frac{2 + 0.5}{4 + 0.5\cdot2},
\]
\[
Q(Y) = \frac{0 + 0.5}{0 + 0.5\cdot2} \cdot \frac{1 + 0.5}{1 + 0.5\cdot2} \cdot \frac{0 + 0.5}{2 + 0.5\cdot2} \cdot \frac{1 + 0.5}{3 + 0.5\cdot2} \cdot \frac{2 + 0.5}{4 + 0.5\cdot2},
\]
\[
Q(X, Y) = \frac{0 + 0.5}{0 + 0.5\cdot4} \cdot \frac{0 + 0.5}{1 + 0.5\cdot4} \cdot \frac{0 + 0.5}{2 + 0.5\cdot4} \cdot \frac{0 + 0.5}{3 + 0.5\cdot4} \cdot \frac{1 + 0.5}{4 + 0.5\cdot4},
\]
\[
\frac{3}{256} = Q(X)  > \frac{Q(X,Y)}{Q(Y)} = Q(X \mid Y) = \frac{1}{90}.
\]
Here, when we consider the optimal parent set for \(X\) in the set \(\{X, Y\}\), \(Y\) is not included in the parent set of \(X\).
\section{Existing work}
The state-of-the-art algorithm progresses through several steps to achieve the final globally optimal solution as shown in Figure 2. \begin{figure}[h!]
    \centering
    \includegraphics[width=0.8\textwidth]{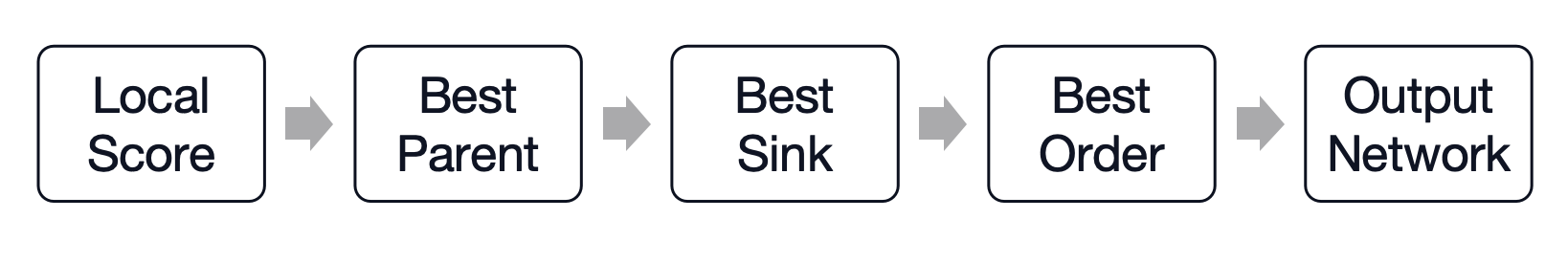}
    \caption{Five essential steps in the state-of-the-art algorithm.}
\end{figure}

The algorithm begins by calculating every possible local score \( Q(S) \) for all subsets \( S \subseteq \{X^{(1)}, \ldots, X^{(p)}\} \). Secondly, for each subset \( S \) and \(X \in S\), the optimal parent set for each node in it is computed by

\begin{equation}
\begin{aligned}
\frac{Q(X,\pi(X,S \backslash X))}{Q(\pi(X,S \backslash X))} &= \max_{T \subseteq S \backslash X} \frac{Q(X,T)}{Q(T)} \\
&= \max \left\{ \frac{Q(X,S \backslash X)}{Q(S \backslash X)}, \max_{Y \in S \backslash X} \frac{Q(X,\pi(X,S \backslash \{X, Y\}))}{Q(\pi(X,S \backslash \{X, Y\}))} \right\},
\end{aligned}
\end{equation}
where \( \pi(X,S \backslash X) \) is the parent set of \(X \in S\) within $S \backslash X$. 

Thirdly, the results from the previous calculations are used for finding the sink nodes. By identifying the sink node for each local structure step by step and then combining the sink nodes corresponding to all variables, the optimal order for global structure can be inferred. For example, for the set \( \{X, S\} \), if \( X \) is known to be the sink variable, and its optimal parent set has been determined in the second step, the information for \( X \) is confirmed for this local structure. Next, only \( S \) needs to be considered.

Fourthly, by utilizing the identified sink nodes to deduce the order. For the set \(\{X, Y, Z\}\), if \(Z\) is the sink node and \(Y\) is the sink node for \(\{X, Y\}\), then \(X\) is the most upstream node and has no parent set. The parent sets of \(Y\) and \(Z\) can be directly obtained from the corresponding results saved in the second step. Step 3 and Step 4 can be summarized by the following equation
\begin{equation}
R(S) = \left\{\begin{array}{ll}
\displaystyle \max_{X \in S} R(S \backslash X) \frac{Q(X,\pi(X,S \backslash X))}{Q(\pi(X,S \backslash X))}, & |S| \geq 1 \\
1, & |S| = 0
\end{array}\right.
.
\end{equation}
Finally, based on the obtained order, by considering the optimal parent set within the corresponding local structure, the optimal Bayesian network can be output.

In the entire process, it is necessary to traverse all combinations three times: first, when calculating the local scores; second, when determining the optimal parent sets; and third, when identifying the optimal sink nodes. It is possible to integrate the first and second steps, meaning that while determining the optimal parent sets, the local scores can also be calculated correspondingly. However, this would still require traversing all combinations at least twice. To address this, we have made improvements in the subsequent proposed method.

\section{Proposed Method}
The core concept of the proposed method is illustrated in Figure 3.
\begin{figure}[h!]
    \centering
    \includegraphics[width=1\textwidth]{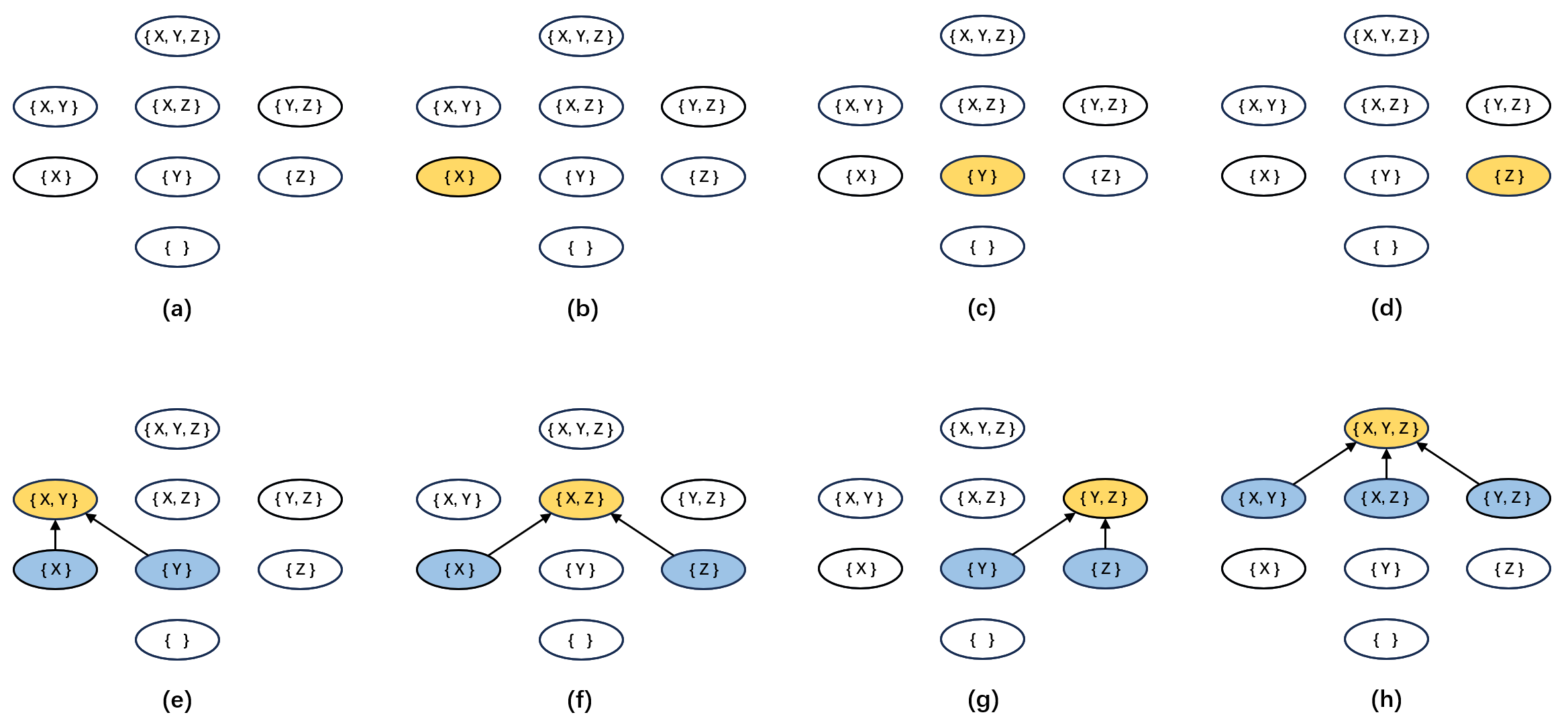}
    \caption{Illustration of the level-by-level computation process in the proposed algorithm. Yellow indicates the combination being computed, while blue represents the data and information required for the computation.}
\end{figure}

Figure 3(a) displays all subsets of the set $\{X, Y, Z\}$. The computation process proceeds level by level as follows:
\begin{itemize}
    \item First Level (Figures 3(b), 3(c), 3(d)): No information is needed from the empty set. Each subset's data is computed individually.
    \item Second Level (Figures 3(e), 3(f), 3(g)): In Figure 3(e), to compute the set \(\{X, Y\}\), data from both \(\{X\}\) and \(\{Y\}\) is required. However, for Figure 3(f), the data from \(\{X\}\) is also needed. Thus, the data saved from the first level is utilized, avoiding redundant computations. During the computation process, the data from the first level is updated and saved into the second level.
\item Third Level (Figure 3(h)): This level involves the complete set of variables. At this stage, only the data from the second level is needed to compute the final results, without referring back to the first level.
\end{itemize}

Next, we will provide the detailed computation process for each set. For every subset \( S \) where $|S| = k$, we assume that \( R(S) \) and \( \pi(X,S\backslash X) \) for each \( X \) in \( S \) have already been computed in the former level, the process is then repeated for each subset \( S \) where $|S|= k+1$. The following is the description of the algorithm steps:
\begin{enumerate}
\item We firstly compute $Q(S)$ for this local structure. 
\item For each $X\in S$, we compute $Q(S\backslash X)$. And we define
$\pi(X,S\backslash X)$ 
by
\begin{equation}
\pi(X,S\backslash X) := \max \left\{ \frac{Q(S)}{Q(S\backslash X)}, \max_{Y \in S \backslash X} \frac{Q(X,\pi(X,S \backslash \{X, Y\}))}{Q(\pi(X,S \backslash \{X, Y\}))} \right\}.
\end{equation}
Here, when \(\frac{Q(S)}{Q(S \setminus X)}\) is maximized, \(\pi(X, S \setminus X)\) retains the corresponding value and sets the optimal parent set of \(X\) in this structure as \(S \setminus X\). When there exists a \(Y \in S \setminus X\) such that \(\max_{Y \in S \setminus X} \frac{Q(X, \pi(X, S \setminus \{X, Y\}))}{Q(\pi(X, S \setminus \{X, Y\}))}\) is the maximum value, \(\pi(X, S \setminus X)\) retains the corresponding value and updates the optimal parent set of \(X\) in this structure with the information from \(\pi(X, S \setminus \{X, Y\})\).

\item Finally, we will find the $X\in S$ that maximizes 
$\displaystyle R(S\backslash X)\frac{Q(X,\pi(X,S\backslash X))}{Q(\pi(X,S\backslash X))}$
to obtain $R(S)$. That is, to find the sink node of \( S \).
\end{enumerate}
In this manner, the optimal structure is consistently obtained for each subset \(S\), with the optimal parent set for each variable continuously updated, ensuring that we ultimately achieve a globally optimal result.

As mentioned above, even if the existing methods combine the steps of calculating the local structure scores and the optimal parent sets, they still require traversing all combinations at least twice. In contrast, the proposed method further integrates the step of calculating the optimal parent sets with the step of identifying the sink nodes for each set. This means that the proposed method only requires traversing all combinations once. This is the primary distinction and improvement over the existing methods. Therefore, even though the time complexity shown in Table 1 is the same, as mentioned, our method traverses all combinations only once. This results in a relatively significant improvement in actual computation time compared to existing work, which is also validated by the results in the experimental section.

For existing work, they claim that the memory complexity of the algorithm is \(O(2^p)\), which is accurate when using a hard disk to store the best parent sets. However, when using only memory, the complexity becomes \(O(p 2^p)\). In this scenario, a computer with 16GB of memory can handle a Bayesian network with a maximum of 26 variables. Through theoretical analysis, we have obtained the following evaluation results. Although the improvement is only by a factor of \(\frac{1}{\sqrt{p}}\), a process that originally required 64GB of memory can now be accomplished with 16GB of memory, highlighting its value. The detailed proof for Table 1 can be found in Appendix A.

\begin{table}[h!]
\centering
\caption{Comparison of computation and memory requirements}
\begin{tabular}{|c|c|c|}
\hline
 & \textbf{Silander-Myllymäki} & \textbf{Proposed method} \\
\hline
Computation & \( O(p^22^p) \) & \( O(p^22^p) \) \\
memory & \( O(p2^p) \) & \(O(\sqrt{p}2^p)\) \\
\hline
\end{tabular}
\label{tab:memory_comparison}
\end{table}

\section{Experiments}
Due to the possibility of exceeding 16GB during the experiments, we used a Core i7-12700K CPU with a 32GB DDR4 memory kit from G.Skill for the relevant experiments in this paper. The algorithm is executed via Rcpp in the R language  (Eddelbuettel, 2013), which can effectively enhance computational efficiency. Considering our exploration of the upper limit of the number of variables, we used the Alarm database with 37 variables (Beinlich et al., 1989). The sample size will not affect the comparison results between the two algorithms, and it is not a limiting factor for Bayesian networks. In the specific experiments, since we mainly compare the time and memory complexity of the two algorithms, we will simply set the sample size to 200.

In Section 5.1, we will present the experimental results comparing the two methods to demonstrate that the proposed method is more advantageous in terms of time and memory complexity. In Section 5.2, we will show the stability of the proposed method when using memory only. In Section 5.3, we will demonstrate the implementation of a Bayesian network with 28 variables using the proposed method with memory only.

\subsection{Comparison of Two Methods}
Based on Table 1 and our calculations, we can determine the maximum number of variables that can be handled by the two methods when using only 16GB of memory. For the existing method, the upper limit is 26 variables, whereas our proposed method can handle up to 28 variables. In the paper by Silander and Myllymäki, they claim that their algorithm is feasible with $p < 33$ when using disk storage and 16GB of memory, and they demonstrated with $p = 29$. However, with $p = 32$, the variables alone already consume 16GB of memory, making it impractical when considering system memory usage and the memory required for software operation.
To maintain consistency, since both the proposed method and the existing method can implement a Bayesian network with 25 variables. Since the memory and time required are minimal for less than 20 variables, our comparative experiments will start from 20 and go up to 25 variables. Each algorithm will be run ten times for each number of variables, and the average will be taken.

As shown in Figure 4(a), the peak memory usage of the proposed method is significantly lower than that of the existing method across different numbers of variables. This validates the superiority of the proposed method in terms of memory usage. As shown in Figure 4(b), even though the theoretical computational complexity is similar, the algorithm is more efficient, resulting in a noticeable improvement in runtime.

\begin{figure}[h!]
    \centering
    \begin{subfigure}[b]{0.45\textwidth}
        \includegraphics[width=\textwidth]{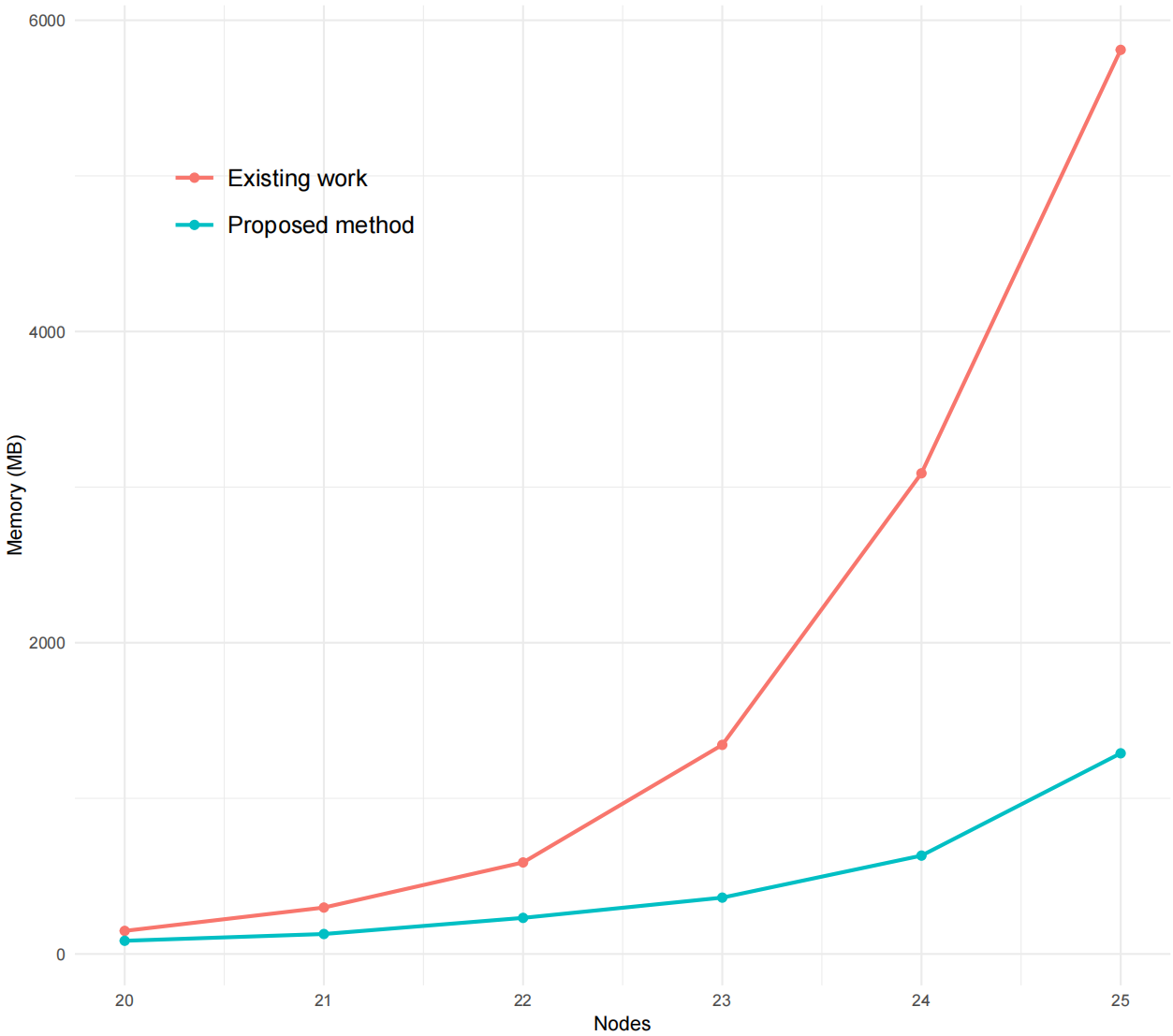}
        \caption{Peak memory usage}
        \label{fig:peak_memory}
    \end{subfigure}
    \hfill
    \begin{subfigure}[b]{0.45\textwidth}
        \includegraphics[width=\textwidth]{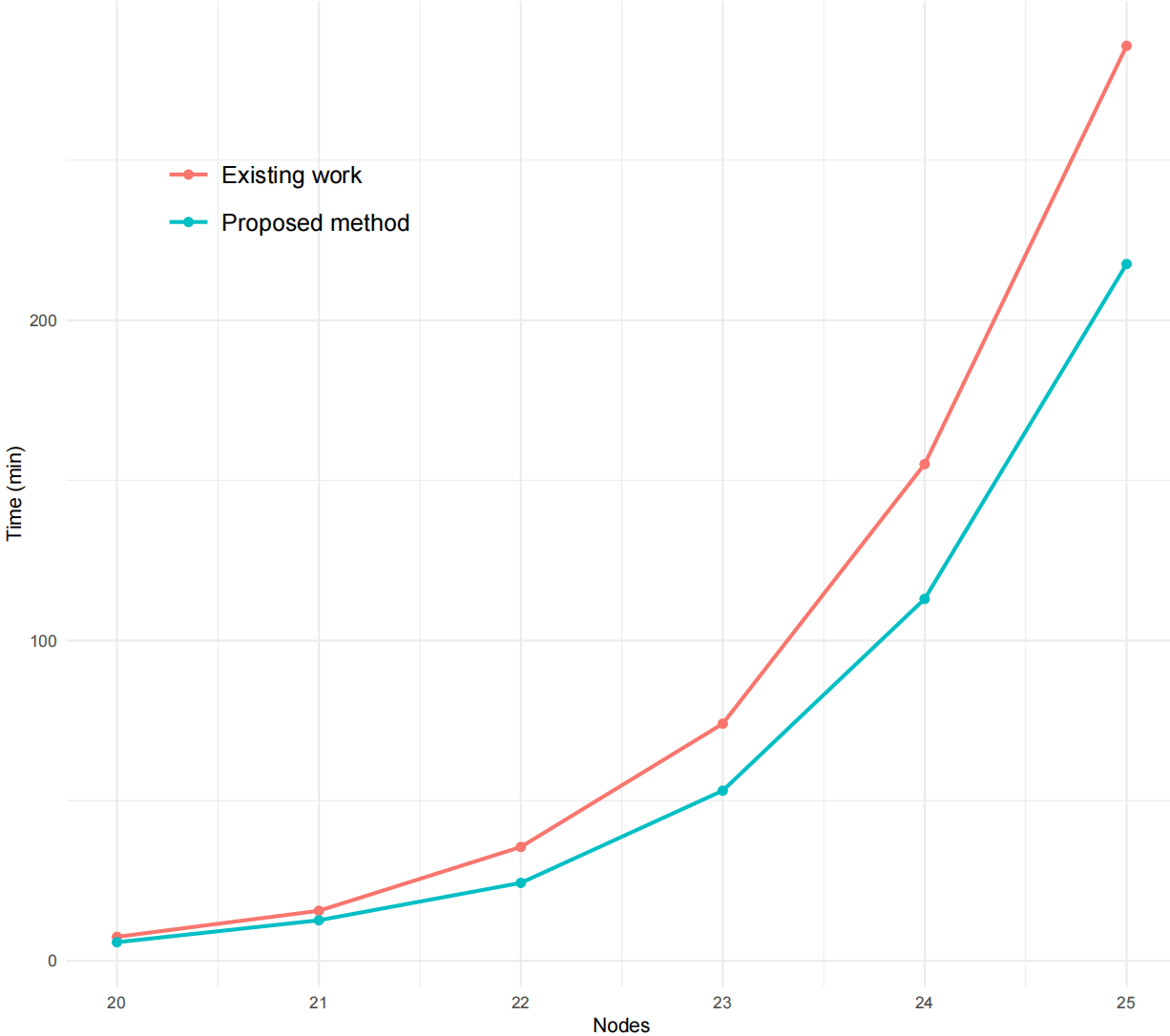}
        \caption{Execution time}
        \label{fig:execution_time}
    \end{subfigure}
    \caption{Comparison of different metrics.}
    \label{fig:comparison}
\end{figure}
The specific data corresponding to different numbers of variables is detailed in Table 2. The table illustrates how the proposed method consistently outperforms the existing method in both memory usage and execution time. This performance advantage becomes increasingly significant as the number of variables increases, highlighting the scalability and robustness of the proposed approach. 

\begin{table}[h!]
\centering
\caption{Detailed data for variables ranging from 20 to 25.}
\resizebox{\textwidth}{!}{%
\begin{tabular}{cccccc}
\toprule
\multirow{2}{*}{Number of Variables} & \multicolumn{2}{c}{Time (min)} & & \multicolumn{2}{c}{Memory (MB)} \\ \cmidrule{2-3} \cmidrule{5-6} 
                        & Existing Work & Proposed Method & & Existing Work & Proposed Method \\ \midrule
20                      & 7.50          & 5.21            & & 148.43        & 84.86            \\ 
21                      & 15.63         & 10.46          & & 298.33        & 128.30           \\ 
22                      & 35.55         & 21.96           & & 588.16        & 232.00           \\ 
23                      & 74.03         & 45.69           & & 1343.69       & 362.12           \\ 
24                      & 155.10        & 99.85          & & 3088.90       & 632.11           \\ 
25                      & 285.67        & 217.70          & & 5809.79       & 1289.59          \\ \bottomrule
\end{tabular}%
}
\label{table:comparison}
\end{table}
\subsection{Stability Test of the Proposed Method}
In this subsection, we will present the experimental results of the proposed method, verifying its stability in terms of time and memory complexity when using only memory. We will present the comparison between the results of ten experiments and their average values. The specific data will be presented in Tables 3 and 4 in Appendix B. For memory complexity, by properly allocating memory resources and effectively managing memory usage, unnecessary memory leaks and overflows can be avoided, thereby enhancing the stability and performance of the program. As shown in Figure 5(a), although there are some fluctuations during code execution, the memory usage remains relatively stable. For runtime, as shown in Figure 5(b), the runtime remains stable regardless of the number of variables.
\begin{figure}[h!]
    \centering
    \begin{subfigure}[b]{0.45\textwidth}
        \includegraphics[width=\textwidth]{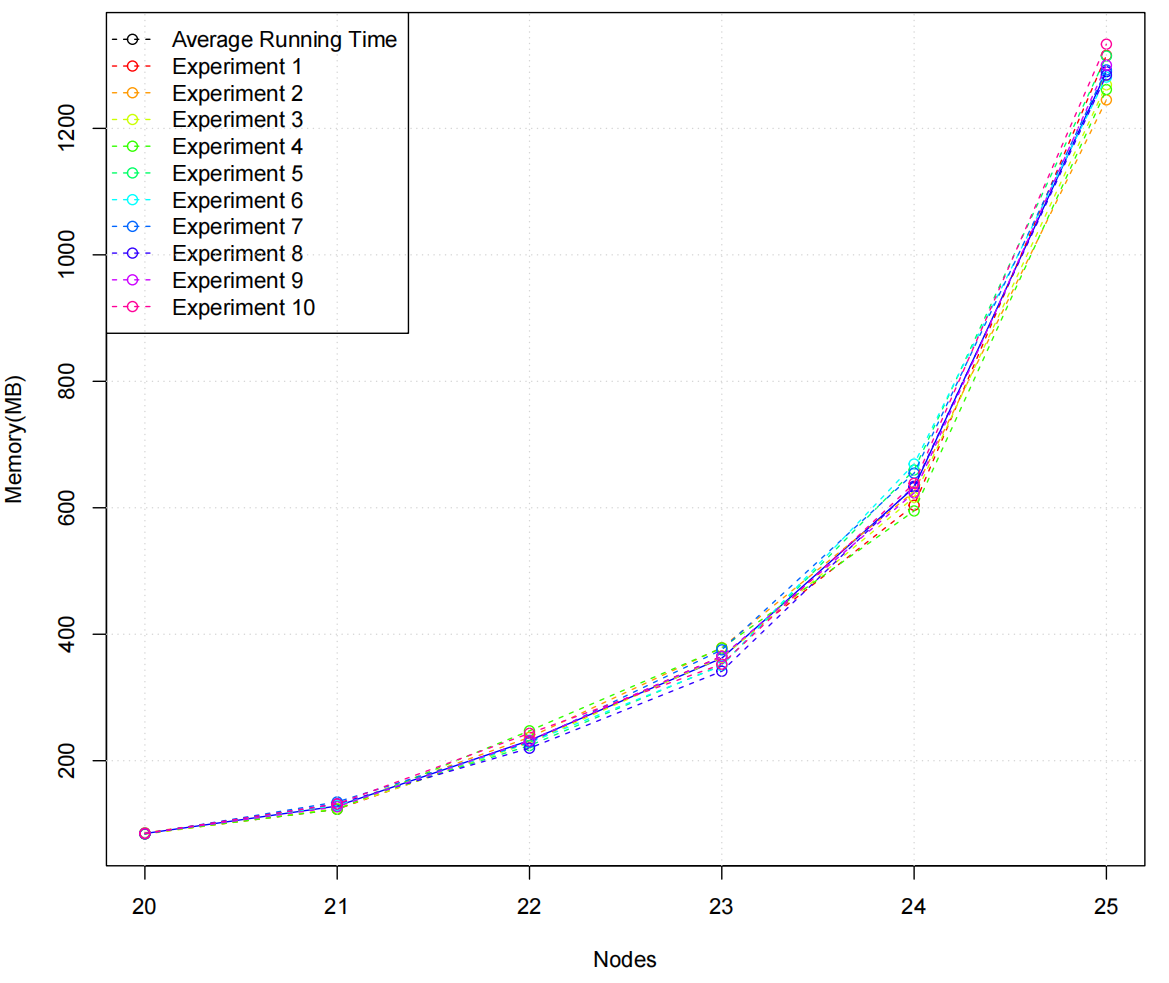}
        \caption{Peak memory usage}
        \label{fig:peak_memory}
    \end{subfigure}
    \hfill
    \begin{subfigure}[b]{0.45\textwidth}
        \includegraphics[width=\textwidth]{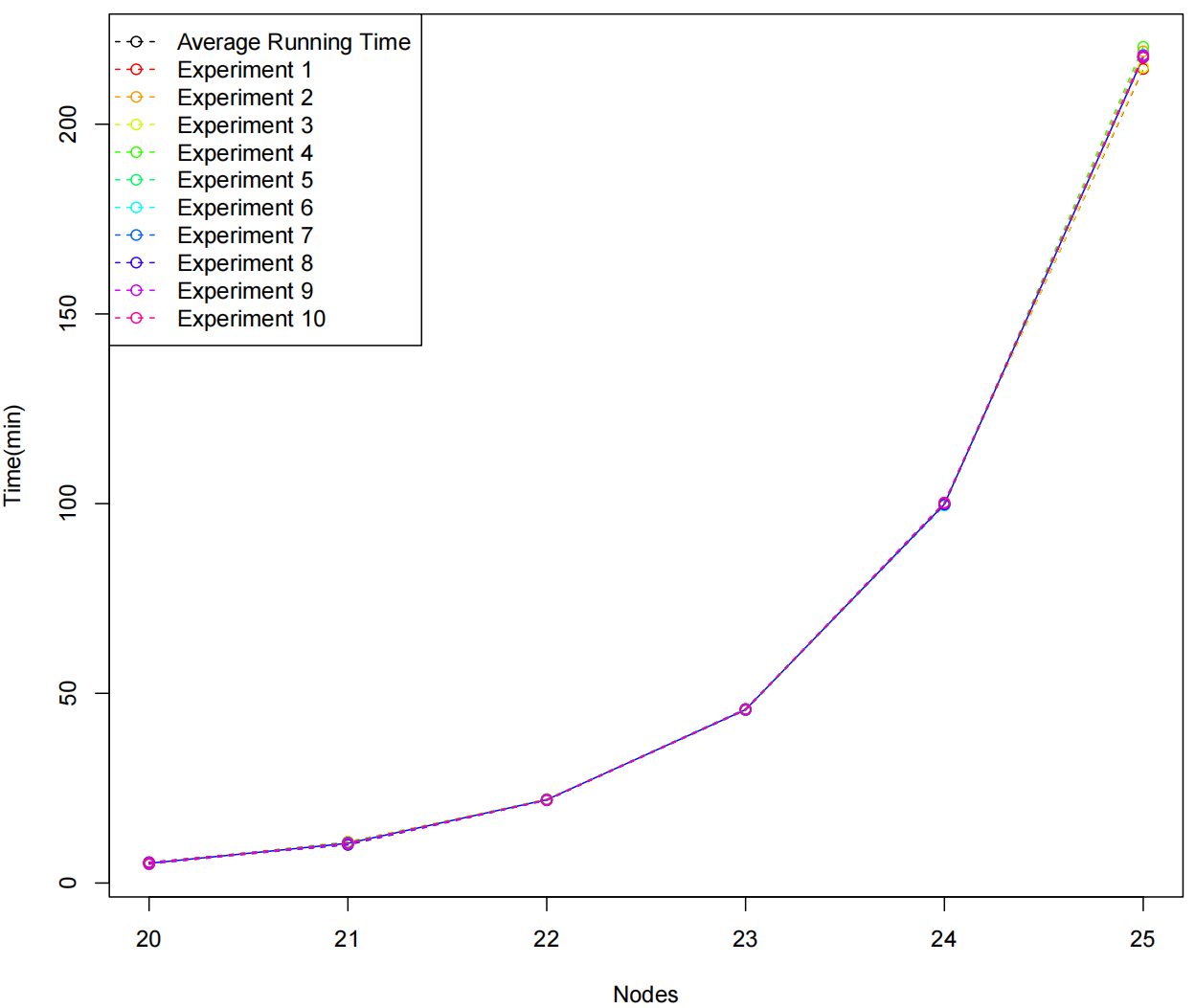}
        \caption{Execution time}
        \label{fig:execution_time}
    \end{subfigure}
    \caption{Verification of the stability of the proposed method.}
    \label{fig:comparison}
\end{figure}

\subsection{Constructing a large network}
Theoretically, the maximum Bayesian network that can be run using only memory is 28 variables, so we chose to implement this. As mentioned in the experimental details of this section, we will use the first 28 variables of the Alarm dataset and a sample size of 200 for computation. The final output results are shown in Figure 6.
\begin{figure}[h!]
    \centering
    \includegraphics[width=0.8\textwidth]{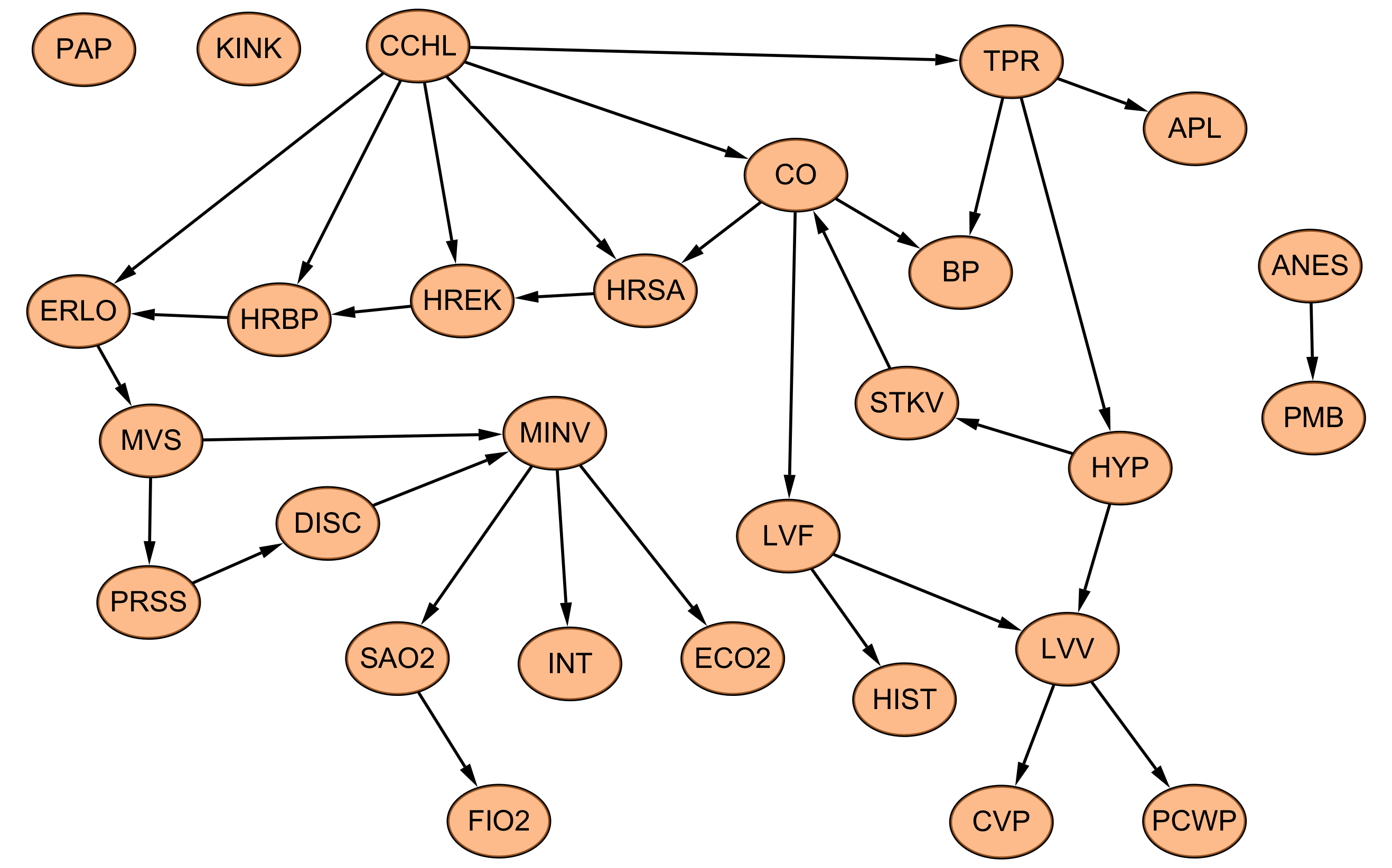}
    \caption{The output of running on the first 28 variables of the Alarm dataset.}
\end{figure}

We used the hardware equipment mentioned at the beginning of this section to run this code. The specific experiment took 1944.43 minutes, and the peak memory usage was 10070.62MB. From the specific data, it can be seen that it is impossible to construct a Bayesian network with 29 variables using only memory. However, for the proposed method, it seems that there are still some interesting points worth analyzing.

As shown in Figure 7, when we solve for the optimal Bayesian network level by level, we are still constrained by the number of combinations. It can be seen that the number of combinations near the middle level is much larger than at both ends. Considering 29 variables, according to \(O(\sqrt{p}2^p)\), we can deduce that the 15th level will be the peak of memory usage. Specifically, the most memory-consuming part is storing the optimal parent set vector for each variable. Although it is the 15th level, the parent set of a variable cannot include itself, so we first need to calculate the combination number \(C(28, 14)\), then multiply by the number of variables, and consider that the stored data is in double format. Finally, considering this vector alone takes up 8.6679GB of memory. Considering that we need to store the data of two levels, it is impossible to construct a Bayesian network with 29 variables.

However, from another perspective, once we store the optimal parent set vector of one level on disk, it becomes quite easy to construct a Bayesian network with 29 variables. Compared to existing work, the proposed method can reduce the memory peak by using the disk only at the peak or near-peak levels, rather than throughout the entire process. For instance, if we use the disk at the 15th level, the vector mentioned above would only occupy 0.2989GB when called. Additionally, compared to existing work, the vectors called from the disk would be shorter in length, making them easier to read and write.
\begin{figure}[h!]
    \centering
    \includegraphics[width=0.8\textwidth]{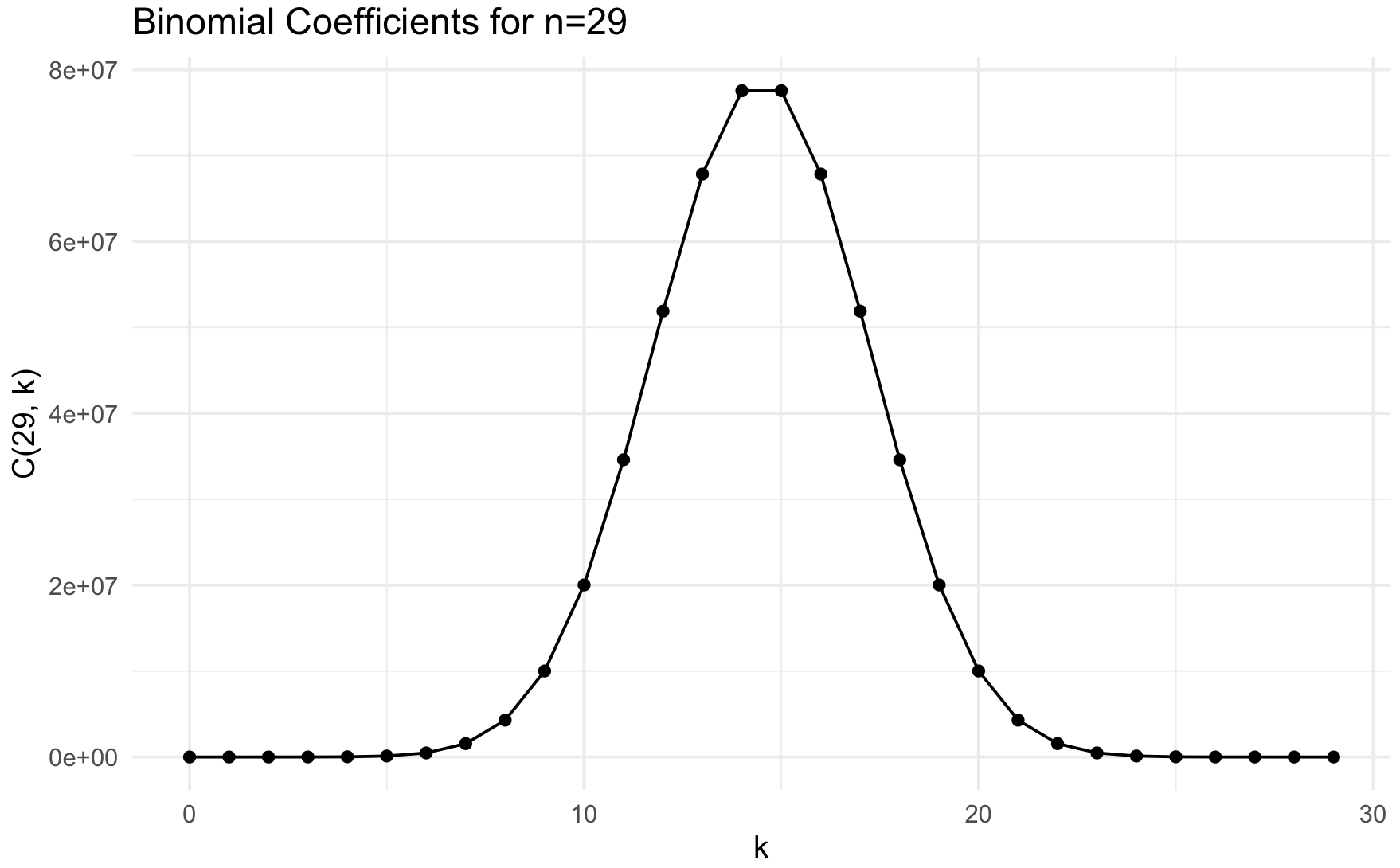}
    \caption{The change in the number of combinations when the number of variables is 29.}
\end{figure}

\section{Conclusion}
In this paper, we propose a more efficient algorithm for solving the global optimal Bayesian network structure. Existing methods use disk storage to accommodate larger vectors for Bayesian networks with more variables, but this approach is prone to disk I/O bottlenecks, leading to a decline in overall runtime efficiency. Therefore, we aim to propose a method that can achieve Bayesian networks with a sufficient number of variables using only memory. Unlike existing methods that proceed step-by-step, we reduce peak memory usage by employing a more flexible, layer-by-layer progression from upstream to downstream in the structure, enabling networks with more variables. Additionally, while existing methods require at least two full iterations over all combinations, our proposed method can achieve the same with just one iteration. Thus, although the theoretical time complexity of our method is consistent with that of existing methods, experimental results demonstrate a significant improvement in runtime efficiency.

Moreover, experiments have validated that our proposed method significantly reduces peak memory usage compared to existing methods when using memory alone, highlighting its substantial advantages in both time and memory complexity. We also implemented a Bayesian network with 28 variables based on the Alarm dataset, further demonstrating the superiority of our method. Specifically, since the memory space occupied by vectors storing the optimal parent sets of variables in the middle layers is significantly larger than in other layers, our method can also further reduce memory complexity if disk storage is needed, without requiring disk usage throughout the entire computation. Disk usage is only necessary when the number of combinations increases significantly, thus avoiding the drawbacks of disk usage. In the future, we will further explore how to improve Bayesian networks in terms of runtime efficiency and peak memory limitations.


\section*{Appendix}
\subsection*{Appendix A}
Assuming there are a total of \( p \) variables, for the \( k \)-th level, where the sets contain \( k \) variables. We here consider time complexity firstly.

1. The first step computes the local score with a time complexity of \( \binom{p}{k} \).

2. In the second step, for each set, we need to update the optimal parent set for each variable. According to Equation (10), the time complexity is
   \[
   \binom{p}{k} \binom{k}{1} \binom{k-1}{1}.
   \]

3. In the third step, using the data from the current and previous layers, we compute the sink node for each set. The time complexity is
   \[
   \binom{p}{k} \binom{k}{1}.
   \]

We aim to compute the complexity of the following sum:

\[
\sum_{k=1}^{p} \left[ \binom{p}{k} + \binom{p}{k} k (k-1) + \binom{p}{k} k \right].
\]

We start by breaking this sum into three parts. First, considering  the binomial theorem, we simply get:
\[
\sum_{k=1}^{p} \binom{p}{k} = 2^p - 1.
\]

Next, we consider the sum:
\[
\sum_{k=1}^{p} \binom{p}{k} k (k-1).
\]

Using the identity \( k(k-1) \binom{p}{k} = p(p-1) \binom{p-2}{k-2} \), the sum becomes:

\[
\sum_{k=1}^{p} k(k-1) \binom{p}{k} = \sum_{k=2}^{p} k(k-1) \binom{p}{k} = p(p-1) \sum_{k=2}^{p} \binom{p-2}{k-2}.
\]

Changing the index of summation:

\[
\sum_{k=2}^{p} \binom{p-2}{k-2} = \sum_{j=0}^{p-2} \binom{p-2}{j} = 2^{p-2}.
\]

Thus, we have:

\[
\sum_{k=1}^{p} k(k-1) \binom{p}{k} = p(p-1) \cdot 2^{p-2}.
\]

Similar to the above derivation process, we can obtain the third part as follows: 

\[
\sum_{k=1}^{p} k \binom{p}{k} = p \sum_{k=1}^{p} \binom{p-1}{k-1} = p \sum_{j=0}^{p-1} \binom{p-1}{j} = p \cdot 2^{p-1}.
\]

Combining all three parts, we obtain:

\[
\sum_{k=1}^{p} \left[ \binom{p}{k} + \binom{p}{k} k (k-1) + \binom{p}{k} k \right] = (2^p - 1) + p(p-1) \cdot 2^{p-2} + p \cdot 2^{p-1}.
\]
Therefore, the time complexity is \( O(p^22^p) \).

Next, we calculate the memory complexity. For binomial coefficients, the middle value \( \binom{p}{k} \) reaches its maximum when \( k \) is approximately \( \frac{p}{2} \). However, we must consider the vector that stores the optimal parent set, which corresponds to the scenario where \( k\frac{p}{k} \) reaches its maximum value. Therefore, we need further analysis.

Stirling's approximation provides an approximate value for \( n! \)

\[
n! \approx \sqrt{2 \pi n} \left( \frac{n}{e} \right)^n.
\]

Using Stirling's approximation, we obtain

\[
k\binom{p}{k} = \frac{k \cdot p!}{k! (p-k)!} \approx \frac{k\sqrt{2 \pi p} \left( \frac{p}{e} \right)^p}{\sqrt{2 \pi k} \left( \frac{k}{e} \right)^k \cdot \sqrt{2 \pi (p-k)} \left( \frac{p-k}{e} \right)^{p-k}}.
\]

After simplification, let \( f(k) = \frac{k \sqrt{2 \pi p}}{2 \pi \sqrt{k(p-k)}} \cdot \frac{p^p}{k^k (p-k)^{p-k}} \). Taking the natural logarithm, we get:

\[
\ln f(k) = \ln \left( \frac{k \sqrt{2 \pi p}}{2 \pi \sqrt{k(p-k)}} \right) + \ln \left( \frac{p^p}{k^k (p-k)^{p-k}} \right).
\]

Then we can Simplify it as 
\[
\ln f(k) = \ln k + \frac{1}{2} \ln (2 \pi p) - \ln (2 \pi) - \frac{1}{2} \ln k - \frac{1}{2} \ln (p - k) + p \ln p - k \ln k - (p-k) \ln (p-k).
\]

Taking the derivative with respect to \( k \):

\[
\frac{d}{dk} \ln f(k) = \frac{1}{k} - \frac{1}{2k} + \frac{1}{2(p-k)} - \ln k - 1 + \ln (p-k) + 1
\]
\[
=  \frac{1}{2k} + \frac{1}{2(p-k)} - \ln k + \ln (p-k).
\]

At this point, we need to find the value of \( k \) such that the following equation holds
\[
\frac{1}{2k} + \frac{1}{2(p-k)} - \ln k + \ln (p-k) = 0.
\]
In fact, it is quite clear that the point at which the maximum value is achieved is slightly greater than \( k = \frac{p}{2} \). However, in our actual situation, we take the value as an integer. Therefore, to obtain the memory complexity, we can simply take \( k = p/2 \) for the calculation.

Using Stirling's approximation, we obtain
\[
\frac{p}{2}\binom{p}{\frac{p}{2}} \approx \frac{p}{2} \cdot \frac{\sqrt{2 \pi p} \left( \frac{p}{e} \right)^p}{\left( \sqrt{\pi p} \left( \frac{p}{2e} \right)^{\frac{p}{2}} \right)^2} = \frac{2^{p} \sqrt{2 \pi p}}{2 \pi}.
\]

It proves that the algorithm's memory complexity is \( O(\sqrt{p}2^p) \).
\subsection*{Appendix B}
\begin{table}[h!]
\centering
\caption{Average Runtime and Experimental Results (in minutes)}
\begin{tabular}{ccccccc}
\toprule
Variables & Avg Time & Exp 1 & Exp 2 & Exp 3 & Exp 4 & Exp 5 \\
\midrule
20 & 5.206 & 4.946 & 5.021 & 5.278 & 5.344 & 5.013 \\
21 & 10.461 & 10.877 & 10.159 & 10.905 & 10.473 & 10.535 \\
22 & 21.965 & 21.899 & 22.051 & 22.083 & 22.110 & 21.958 \\
23 & 45.692 & 45.587 & 45.642 & 45.629 & 45.620 & 45.609 \\
24 & 99.847 & 100.245 & 99.679 & 99.925 & 99.506 & 99.692 \\
25 & 217.699 & 214.567 & 219.227 & 215.028 & 220.421 & 218.180 \\
\midrule
Variables &  & Exp 6 & Exp 7 & Exp 8 & Exp 9 & Exp 10 \\
\midrule
20 &  & 5.269 & 5.312 & 5.492 & 5.005 & 5.382 \\
21 &  & 10.127 & 10.675 & 10.051 & 10.292 & 10.514 \\
22 &  & 22.058 & 21.826 & 21.863 & 21.807 & 21.992 \\
23 &  & 45.928 & 45.623 & 45.814 & 45.643 & 45.830 \\
24 &  & 99.539 & 99.791 & 99.767 & 100.079 & 100.252 \\
25 &  & 218.271 & 217.840 & 217.612 & 218.195 & 217.649 \\
\bottomrule
\end{tabular}
\end{table}

\begin{table}[h!]
\centering
\caption{Average Peak Memory Usage and Experimental Results (in MB)}
\begin{tabular}{ccccccc}
\toprule
Variables & Avg Peak Memory & Exp 1 & Exp 2 & Exp 3 & Exp 4 & Exp 5 \\
\midrule
20 & 84.857 & 84.814 & 86.135 & 83.804 & 85.110 & 84.583 \\
21 & 128.304 & 123.250 & 128.234 & 124.374 & 123.120 & 128.951 \\
22 & 231.996 & 230.666 & 236.945 & 228.125 & 247.300 & 223.971 \\
23 & 362.122 & 365.414 & 378.747 & 363.225 & 378.199 & 351.255 \\
24 & 632.109 & 603.989 & 625.571 & 616.573 & 595.036 & 659.739 \\
25 & 1289.586 & 1315.434 & 1245.000 & 1268.855 & 1261.305 & 1314.414 \\
\midrule
Variables &  & Exp 6 & Exp 7 & Exp 8 & Exp 9 & Exp 10 \\
\midrule
20 &  & 85.185 & 84.274 & 84.645 & 85.058 & 84.960 \\
21 &  & 129.622 & 134.812 & 131.821 & 127.125 & 131.727 \\
22 &  & 228.703 & 229.264 & 219.751 & 231.889 & 243.350 \\
23 &  & 350.183 & 375.353 & 341.457 & 364.786 & 352.606 \\
24 &  & 669.205 & 654.691 & 634.061 & 623.596 & 638.635 \\
25 &  & 1281.074 & 1292.387 & 1284.350 & 1299.879 & 1333.159 \\
\bottomrule
\end{tabular}
\end{table}

\end{document}